\documentclass[10pt,twocolumn,letterpaper]{article}

\usepackage{cvpr}              % To produce the CAMERA-READY version
\usepackage{graphicx}
\usepackage{amsmath}
\usepackage{amssymb}
\usepackage{booktabs}

\usepackage[pagebackref,breaklinks,colorlinks]{hyperref}

\usepackage[capitalize]{cleveref}
\crefname{section}{Sec.}{Secs.}
\Crefname{section}{Section}{Sections}
\Crefname{table}{Table}{Tables}
\crefname{table}{Tab.}{Tabs.}

\usepackage{algorithm}
\usepackage{algpseudocode}
\usepackage{xcolor}
\usepackage{hhline}
\usepackage{array}
\usepackage{multirow}
\usepackage{enumitem}
\newcolumntype{P}[1]{>{\centering\arraybackslash}p{#1}}
\newcolumntype{M}[1]{>{\centering\arraybackslash}m{#1}}
\newcommand{\PreserveBackslash}[1]{\let\temp=\\#1\let\\=\temp}
\newcolumntype{C}[1]{>{\PreserveBackslash\centering}p{#1}}
\newcolumntype{R}[1]{>{\PreserveBackslash\raggedleft}p{#1}}
\newcolumntype{L}[1]{>{\PreserveBackslash\raggedright}p{#1}}

\newcommand*{\val}[2]{{#1}\textcolor{black!75}{\tiny$\pm$#2}}
\newcommand*{\bval}[2]{{\bf #1}\textcolor{black!75}{\tiny$\pm$#2}}

\usepackage{tikz}
\usetikzlibrary{tikzmark}

\def\cvprPaperID{*****} % *** Enter the CVPR Paper ID here
\def\confName{CVPR}
\def\confYear{2023}

\begin{document}

\title{Data-Free Knowledge Distillation Using Adversarially\\Perturbed OpenGL Shader Images}

\author{Logan Frank \hspace{2cm} Jim Davis \\
Department of Computer Science and Engineering \\
Ohio State University \\
{\tt\small \{frank.580, davis.1719\}@osu.edu}
}
\maketitle

%%%%%%%%% ABSTRACT
\begin{abstract}

Knowledge distillation (KD) has been a popular and effective method for model compression. One important assumption of KD is that the original training dataset is always available. However, this is not always the case due to privacy concerns and more. In recent years, ``data-free" KD has emerged as a growing research topic which focuses on the scenario of performing KD when no data is provided. Many methods rely on a generator network to synthesize examples for distillation (which can be difficult to train) and can frequently produce images that are visually similar to the original dataset, which raises questions surrounding whether privacy is completely preserved. In this work, we propose a new approach to data-free KD that utilizes unnatural OpenGL images, combined with large amounts of data augmentation and adversarial attacks, to train a student network. We demonstrate that our approach achieves state-of-the-art results for a variety of datasets/networks and is more stable than existing generator-based data-free KD methods. Source code will be available in the future.
\end{abstract}

%%%%%%%%% BODY TEXT
\section{Introduction} \label{sec:introduction}

Neural networks have become a dominant force in machine learning (ML) \cite{He2016a, Vaswani2017a, Dosovitskiy2020a}. Enabled by modern compute hardware, the size of the largest available networks has continued to increase with time \cite{Kolesnikov2020a, Dehghani2023a, Kirillov2023a}. Coinciding with the desire to increase model capacity, there has been a growing interest in deploying small, but well-performing, networks on edge devices \cite{Wang2022a, Singh2023a}. Utilizing neural networks on such devices is often difficult as there are typically strict resource constraints such as memory or power consumption. Unfortunately, models frequently have to sacrifice performance in order to satisfy these imposed limitations. This often points practitioners towards considering existing methods that enable a smaller network to perform as well as a larger network all while retaining its lean characteristic. 

Paradigms that address the aforementioned goal are called model compression techniques, which as the name implies, aim to compress a large and complex neural network that performs well on some downstream task (classification, object detection, etc.) into one with a more compact and efficient form. One such method is \textit{knowledge distillation} (KD) \cite{Hinton2015a}, where the goal is to transfer the information stored inside a cumbersome ``teacher" network to a more compact ``student" network (which can be a completely different architecture). This is accomplished by utilizing the soft-target outputs (or internal features) of the teacher network to guide the student towards mimicking the output responses of the teacher when presented with similar inputs. It has even been shown that performance gains can be achieved by distilling between a teacher and student that are the exact same network architecture (\textit{self-distillation}) \cite{Furlanello2018a}. Other works have built upon the standard KD approach by applying adversarial attacks to the original data for exploiting decision boundary information in the teacher model, which resulted in improved student accuracy \cite{Heo2019a, Tian2023a}. 

\begin{figure}[t]
\begin{center}
\begin{subfigure}{.105\textwidth}
\includegraphics[width=1.7cm]{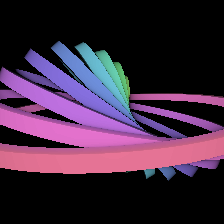}
\end{subfigure}
\begin{subfigure}{.105\textwidth}
\includegraphics[width=1.7cm]{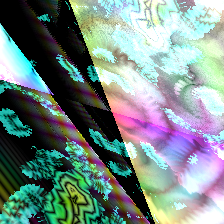}
\end{subfigure}
\begin{subfigure}{.105\textwidth}
\includegraphics[width=1.7cm]{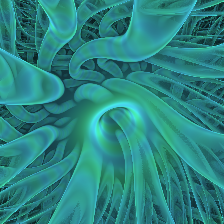}
\end{subfigure}
\begin{subfigure}{.105\textwidth}
\includegraphics[width=1.7cm]{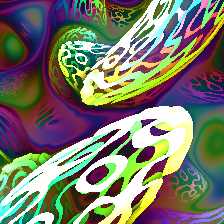}
\end{subfigure}
\\ \vspace{0.1cm}
\begin{subfigure}{.105\textwidth}
\includegraphics[width=1.7cm]{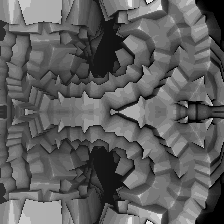}
\end{subfigure}
\begin{subfigure}{.105\textwidth}
\includegraphics[width=1.7cm]{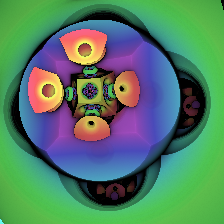}
\end{subfigure}
\begin{subfigure}{.105\textwidth}
\includegraphics[width=1.7cm]{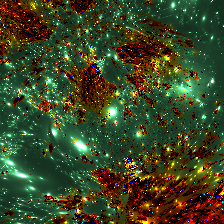}
\end{subfigure}
\begin{subfigure}{.105\textwidth}
\includegraphics[width=1.7cm]{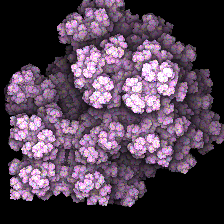}
\end{subfigure}
\end{center}
\vspace{-0.2in}
\caption{Example TwiGL OpenGL shader images.}
\label{fig:synthetic_images}
\vspace{-0.14in}
\end{figure}

% DFKD paragraph
While KD has been a remarkably popular mechanism for creating lightweight models, one particular drawback that it (along with other model compression techniques \cite{Janowsky1989a, Gong2014a} and KD derivatives \cite{Heo2019a, Tian2023a}) suffers from is that standard KD relies on the strong assumption that the original training data (or similar distribution data) will be available at distillation. However, this is not often the case as privacy concerns continue to grow regarding the use and transferring of data. Moreover, the data could be kept as proprietary in-house information by a company or simply put, the dataset is too large to transfer or store in a reasonable manner. To circumvent this drawback, many approaches have been proposed in the realm of \textit{data-free knowledge distillation} (DFKD), which utilizes the teacher as a means to generate synthetic examples to orchestrate the distillation process. These methods typically involve generating examples by either sampling from distributions modeled off prior information contained inherently in the teacher (\eg, batch norm \cite{Ioffe2015a} statistics) \cite{Nayak2019a, Wang2021a} or an additional generator network  \cite{Goodfellow2014a} trained concurrently with distillation \cite{Chen2019a, Choi2020a, Binici2022a}. Incidentally, the synthetic examples generated by these approaches often closely resemble or give insights to the original data \cite{Yin2020a, Yoo2019a}. This begs the question as to whether or not these DFKD approaches are truly privacy-preserving. Furthermore, the potential inclusion of a generator network can complicate the KD process as there is a completely different set of hyperparameters to be managed as well as the potential issues of mode collapse or non-convergence \cite{Saxena2021a}.

% We additionally show qualitatively that our distilled models are more robust to model inversion
In this work, we present a novel approach to DFKD that differs from existing methods in that it does not utilize an additional generator network or attempt to extract naturalistic synthetic examples from the teacher. Instead, we propose to 1) construct a dataset of unnatural synthetic images using OpenGL shaders, 2) adversarially perturb them to identify the decision boundaries in the teacher model, and 3) transfer this knowledge to the student using standard KD \cite{Hinton2015a}. Experiments show significant improvements over existing DFKD approaches for multiple datasets and network architectures, all while being completely privacy-preserving with respect to the original dataset. We additionally show qualitatively that our distilled models can better anonymize the original data embedded in the teacher. Our contributions are summarized as follows:
% We will demonstrate our proposed approach through self-distillation and knowledge distillation scenarios on multiple datasets and network architectures. Additionally, we will conduct a qualitative privacy study on our distilled student models.
\begin{enumerate}[noitemsep,nolistsep]
    \item A new framework for data-free knowledge distillation that exploits synthetic imagery and adversarial perturbations rather than utilizing a generator network.
    \item The proposed approach utilizes the standard knowledge distillation regime and can easily be incorporated into more advanced distillation techniques.
    \item Our method is completely privacy-preserving to the original dataset and furthermore enables a path for anonymizing the data embedded in the teacher model.
\end{enumerate}

\noindent We begin with a review of related work in Sect.~\ref{sec:related_work}. The components of our proposed DFKD approach are described in Sect.~\ref{sec:method}. Lastly, extensive experiments demonstrating our method are presented in Sect.~\ref{sec:experiments}.

\section{Related Work} \label{sec:related_work}

In recent years, many works have been proposed in the areas of KD, DFKD, ML privacy, and adversarial examples.

\smallskip
\noindent \textbf{Knowledge Distillation.} The transferring of knowledge from a large network to a smaller network was introduced in \cite{Bucilua2006a} and was further refined and coined as ``knowledge distillation" in \cite{Hinton2015a}. The general KD framework outlined in \cite{Hinton2015a} trains a student network to match the temperature-scaled \cite{Guo2017a} soft outputs from a larger teacher network using entropy-based loss functions. Since then, several works have investigated what properties influence the success of KD \cite{Urban2017a, Cho2019a, Beyer2022a} and others have proposed structural improvements to the seminal approach \cite{Romero2015a, Shen2020a, Tian2020a, Tarvainen2017a}. Notably, \cite{Beyer2022a} argued that KD can be viewed as ``function matching" and showed that applying mixup \cite{Zhang2018a} to the input distillation images results in improved student performance. They also observed that some knowledge can be transferred to a student network using out-of-domain data, albeit at a significant performance loss compared to employing in-domain data. Furthermore, KD has been coupled with other model compression techniques \cite{Mishra2018a} and applied to other areas such as semi-supervised learning \cite{Xie2020a}, multi-exit architectures \cite{Phuong2019a}, and more. Rather than distilling from a large cumbersome network to a smaller network, self-distillation repeatedly transfers knowledge to a student that is the same architecture as the teacher \cite{Furlanello2018a}. 

\smallskip
\noindent \textbf{Data-Free Knowledge Distillation.} In standard KD, it is assumed that the original training dataset is available. However this is not always the case, which has motivated a series of data-free KD approaches that attempt to transfer knowledge when no data is available. This line of work can be separated into two categories based on whether they utilize a generator network \cite{Chen2019a, Yoo2019a, Choi2020a, Fang2021a, Binici2022a} or inherent teacher network statistics \cite{Nayak2019a, Wang2021a, Yin2020a} to synthesize examples that may be beneficial for KD. The first truly data-free approach was proposed in \cite{Nayak2019a}, where they modelled Dirichlet distributions at the output of the teacher and synthesized examples to match these distributions using backpropagation. In \cite{Chen2019a}, the teacher is treated as a fixed discriminator while a generator is trained to output images that maximize the one-hot cross entropy loss for the teacher. A similar concept is proposed in \cite{Choi2020a} with an added constraint that minimizes the divergence between the generated example's features and the teacher's learned batch norm layer \cite{Ioffe2015a} statistics. % A contrastive loss is introduced in \cite{Fang2021a} to encourage the generator network to synthesize new and novel examples. In addition, they include a ``memory bank" to prevent catastrophic forgetting of examples from earlier in training. In \cite{Binici2022a}, a variational autoencoder \cite{Kingma2013a} is trained to remember previous examples alongside a generator that is trained to produce novel examples. 

Later, we will compare to the relevant approaches of Contrastive Model Inversion (CMI) \cite{Fang2021a} and Pseudo Replay Enhanced DFKD (PRE) \cite{Binici2022a}. Both CMI and PRE rely on a generator network to synthesize examples for transferring knowledge from teacher to student. More specifically, CMI utilizes the framework presented in \cite{Choi2020a}, which consists of three different loss components, with accompanied hyperparameter weight coefficients, that guide the generator's outputs. They introduced a novel contrastive loss component in addition to the aforementioned losses that forces the generator to output new examples that differ from previously synthesized examples that are stored in a memory bank. As for PRE, they utilize the approach outlined in \cite{Chen2019a}, consisting of four loss terms (two of which are similar to CMI), also associated with multiple hyperparameter weight coefficients, used to train the generator network. Rather than using a memory bank, PRE utilized an additional variational autoencoder \cite{Kingma2013a} to remember previously synthesized examples created by the novel-view generator network.

Our work differs significantly from previous methods in that we do not utilize an additional generator network nor do we attempt to synthesize naturalistic examples using the teacher. We do leverage the teacher to obtain information about its decision boundaries, however the samples we obtain remain unnatural in appearance.

\smallskip
\noindent \textbf{ML Privacy.} In recent years, data privacy and the inherent privacy of ML models have become major concerns. It has been shown that private training examples can be extracted from MLP networks \cite{Haim2022a}, large language models \cite{Carlini2021a}, and diffusion models \cite{Carlini2023a}. Although DFKD is motivated by privacy issues surrounding the original dataset, many approaches can generate examples that are visually similar or have artifacts similar to examples from the original data \cite{Yin2020a, Chen2019a, Yoo2019a, Choi2020a, Fang2021a}. This raises an argument as to whether privacy is completely preserved in these methods.

\smallskip
\noindent \textbf{Adversarial Examples.} Another major concern for neural networks is adversarial examples \cite{Szegedy2013a}. These are examples that have been intentionally perturbed to appear unchanged, but have been manipulated by an adversarial attack to be classified by a model as something different (\eg, an image of a ``dog" that was perturbed and classified as ``guacamole"). Numerous adversarial attacks have been proposed with varying benefits and capabilities \cite{Szegedy2013a, Kurakin2016a, Madry2018a, Carlini2017a}. To combat these attacks, several defenses \cite{Papernot2016a, Xie2019a, Raff2019a} and detection \cite{Lust2020a, Abusnaina2021a} methods have been introduced over the years. Rather than attempting to be robust against or detect such inputs, adversarial examples were shown to provide accuracy improvements for standard KD in \cite{Heo2019a, Tian2023a}. These approaches used adversarial attacks on \textit{real} data to create slightly modified examples that helped identify the decision boundaries in the teacher network. We employ adversarial attacks for a similar goal, however we utilize pairs of adversarial examples to better outline the decision boundary and also include an additional stronger/deeper attack. Furthermore, only standard KD with the original \textit{in-domain} training dataset is considered in \cite{Heo2019a, Tian2023a}, whereas we utilize synthetic \textit{out-of-domain} examples in a different KD regime where access to the original data is prohibited.

\section{Method} \label{sec:method}

In this section, we describe the main components of our approach: 1) the creation of a synthetic dataset, 2) the role of data augmentation on that dataset, 3) how to utilize adversarial perturbations on the dataset to identify the decision boundaries of the teacher, and lastly, 4) the method for transferring knowledge from teacher to student.

\subsection{OpenGL Shader Image Dataset}

Our focus for this work is to transfer knowledge from a pretrained network to a freshly initialized network \textit{without} accessing the original dataset. We instead want to create some new dataset with the added constraints that there is \textit{no} additional generator network used for synthesizing these images nor any other mechanism for attempting to recover ``desirable" images from the teacher. Therefore, we leverage procedural image programs (using OpenGL \cite{Shreiner2009a}) for rendering synthetic images to construct the dataset we will use for KD. Utilizing the approach of \cite{Baradad2022a}, we first synthesize several images for each of the available 1089 TwiGL shaders \cite{Twigl2023a}. Examples of these shader images are shown in Fig.~\ref{fig:synthetic_images}. As some of these shaders produced images that were either constant (\ie, containing all one color such as all black or all white) or simple (\ie, containing only two colors or containing few colored pixels), we filter the rendered shader images in order create an initial set of synthetic images with the most diversity possible by removing the aforementioned images. With this scheme, we can synthesize near-infinite amounts of unnatural, \textit{out-of-domain} images to construct a dataset that can be leveraged for KD as will be shown.

Given a teacher network $\mathcal{F}_t$ pretrained on some dataset with $\mathcal{C} = \{ c_1, ..., c_R \}$ classes and a filtered synthetic dataset $\mathcal{D}_S$, we pass every example in $\mathcal{D}_S$ through $\mathcal{F}_t$ to obtain an initial teacher prediction and aggregate all predictions. To form the final synthetic dataset $\mathcal{D}_K$ that will be used for KD, we randomly select examples from $\mathcal{D}_S$ based on their teacher predictions. For each class $c_i$ in the teacher's dataset, we uniformly sample examples from $\mathcal{C}_{-i}$ (the set of classes $\mathcal{C}$ without $c_i$) to meet a desired number of synthetic images $N_i$ per $c_i$ where
% https://mathoverflow.net/questions/49286/notation-for-finite-sequence-with-one-element-is-removed
\begin{align}
    \mathcal{C}_{-i} &= \{\ c_j\ \ \forall\ \ j \in \mathcal{C}\ \ \ \ s.t.\ \ \ \ j \neq i\ \}
\end{align}

\noindent and the set of examples $X_j$ assigned to each $c_j \in \mathcal{C}_{-i}$ are 
\begin{align}
    X_j &= \{\ x_k\ \ \forall\ \ x_k \in \mathcal{D}_S\ \ \ \ s.t.\ \ \ \ \mathcal{F}_t(x_k) = c_j\ \}
\end{align}

\noindent In other words, we are selecting examples for $c_i$ that are predicted as any other label besides $c_i$. If uniformly sampling across $\mathcal{C}_{-i}$ does not meet the specified number of examples per class (\ie, $N_i$ is not evenly divisible by $R-1$), then the remaining examples are randomly sampled without replacement until $N_i$ examples are obtained. For all images collected for class $c_i$, we assign them an associated target label $t = c_i$. Note that we do allow repeats between classes (\ie, a specific image can be assigned to both $c_1$ and $c_2$).

Collecting $N_i$ shader images per class for the specific pretrained teacher network creates the base synthetic distillation dataset $\mathcal{D}_K = \{ (x_1, t_1), ..., (x_N, t_N) \}$ where
\begin{align}
    \mathcal{F}_t(x_j) \neq t_j\ \ \forall\ \ (x_j, t_j) \in \mathcal{D}_K
\end{align}

\noindent Later, we will discuss why it is important that all examples not be classified as their associated target label by default, and further describe how $\mathcal{D}_K$ is used for distilling knowledge from the associated pretrained teacher network to a new, freshly initialized student network. 

\subsection{Data Augmentation} \label{sec:augmentation}

As mentioned, one of the benefits to our method for creating a synthetic dataset is the ability to render an unbounded number of images per shader code, enabling near-infinite dataset sizes if desired. However, it becomes obvious that obtaining hundreds of thousands to potentially millions of images quickly poses a storage issue. A clear solution for increasing dataset diversity without having an over-inflated dataset is data augmentation. This allows us to artificially create more examples during distillation and furthermore, augmented examples could enable us to explore regions of the teacher's feature space that the original data could not discover on its own.
% Furthermore, more shader codes can be included in addition to the TwiGL ones we employed, which could also increase the diversity of the dataset. Having both more images and greater image diversity allows us to sufficiently sample the feature space of the teacher model (likely leading to better knowledge transfer).

In ordinary fully-supervised training, data augmentations should be ``label-preserving" \cite{Geiping2023a}. However, we do not have such constraint as there is no notion of a ``label" (or any other semantic meaning) with our OpenGL shader images. Thus, we can employ large amounts of data augmentation that would normally not be considered. Later, we will describe our complete data augmentation regime for KD training and additionally show how data augmentation has provided a strong benefit to our approach that would likely not be obtained through more physical examples.

\subsection{Decision Boundary Exploitation} \label{sec:adversarial}

To transfer the most amount of information possible to the student, we utilize targeted adversarial attacks on our synthetic examples to identify the decision boundaries in the teacher network. As will be described below, our attack is based on the Basic Iterative Method (BIM) \cite{Kurakin2016a}.

Beginning with a synthetic example $x_j$ having target label $t_j$ where $\mathcal{F}_t(x_j) \neq t_j$, the goal of our adversarial attack is to perturb the image ``from the outside in", starting at an arbitrary label and ending at a desired label $t_j$. This allows us to identify the decision boundary between $t_j$ and all other classes in the teacher $\mathcal{F}_t$. Thus, it is important that the augmented synthetic example be initially classified by the teacher as any other label besides $t_j$. Generally speaking, our attack will perturb an example until it has crossed into the decision space of $t_j$ and is within some specified argmax softmax threshold (to ensure the example is somewhat close to the boundary). In addition to the final adversarial image obtained from a successful attack, we also take the example from the iteration just before the final image crossed the boundary. This creates a pair of examples consisting of the ``post"-success example (the final image) and ``pre"-success example (the second-to-last image). This pre-success image is also required to be within the specified argmax softmax threshold, but be classified as anything besides $t_j$. 

After conducting our adversarial attack on a batch of synthetic images, we filter (remove) examples that do not meet either of the following criteria: 1) image not classified as its assigned target at the end of the adversarial attack or 2) either the final (post-success) image or the pre-success image were unable to meet the softmax threshold. This is to ensure that the examples we are using to guide distillation are beneficial in identifying some meaningful part of the feature space in the teacher, such as the decision boundary.

To aid in the adversarial attack process, we propose adding a Bold Driver heuristic to the attack step size to make it more adaptive and increase its ability to create adversarial examples that meet our conditions. Thus, if after a particular attack iteration the example has not reached the desired target label, the step size will be increased. If the example has crossed the decision boundary to the desired label but the argmax softmax value is above the specific threshold, the example will be reverted back to its previous iteration and the step size will be decreased. Finally, if an example has met the desired final conditions (correct classification label and below argmax softmax threshold), but the pre-success counterpart does not meet the softmax threshold condition, the image will be reverted to the previous iteration and the step size will be decreased (to move the pre-success example closer to the decision boundary).

\begin{figure}[t]
\begin{center}
% \begin{subfigure}{.22\textwidth}
% \includegraphics[width=3.85cm]{images/aadv1.png}
% \caption{}
% \end{subfigure}
\begin{subfigure}{.155\textwidth}
\includegraphics[width=2.67cm]{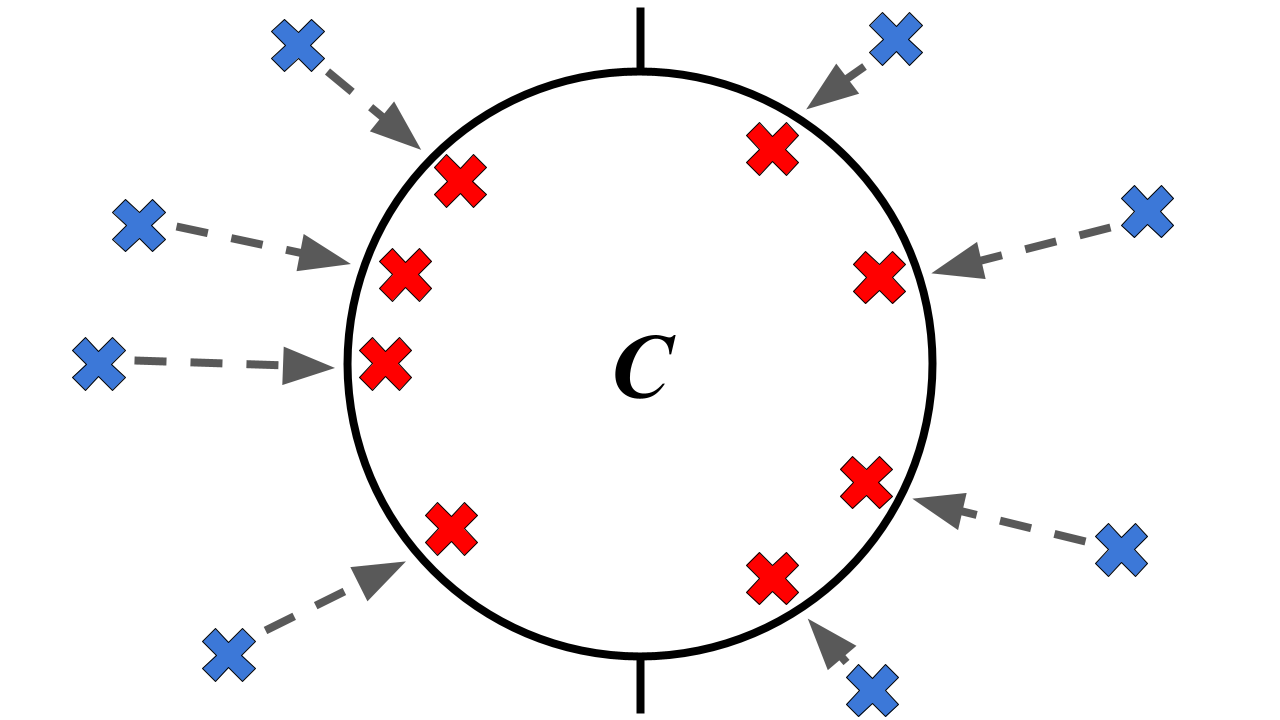}
\caption{}
\end{subfigure} % \\ \vspace{0.3cm}
\begin{subfigure}{.155\textwidth}
\includegraphics[width=2.67cm]{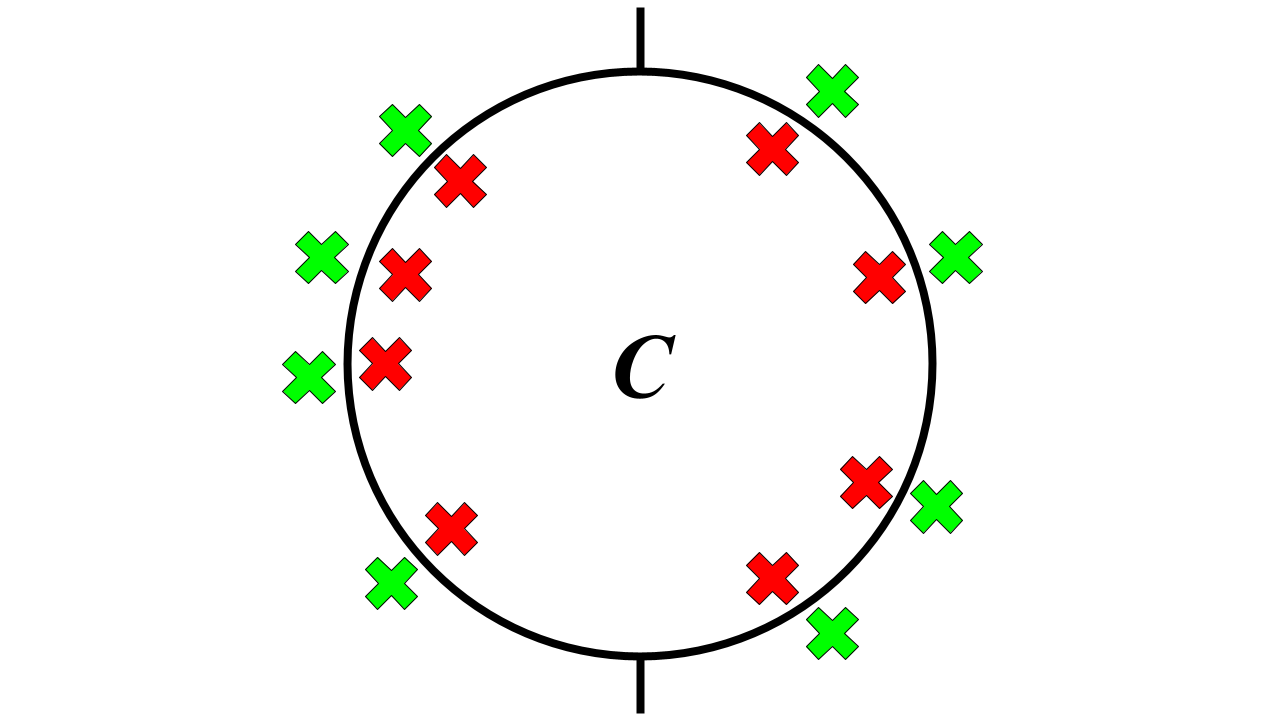}
\caption{}
\end{subfigure}
\begin{subfigure}{.155\textwidth}
\includegraphics[width=2.67cm]{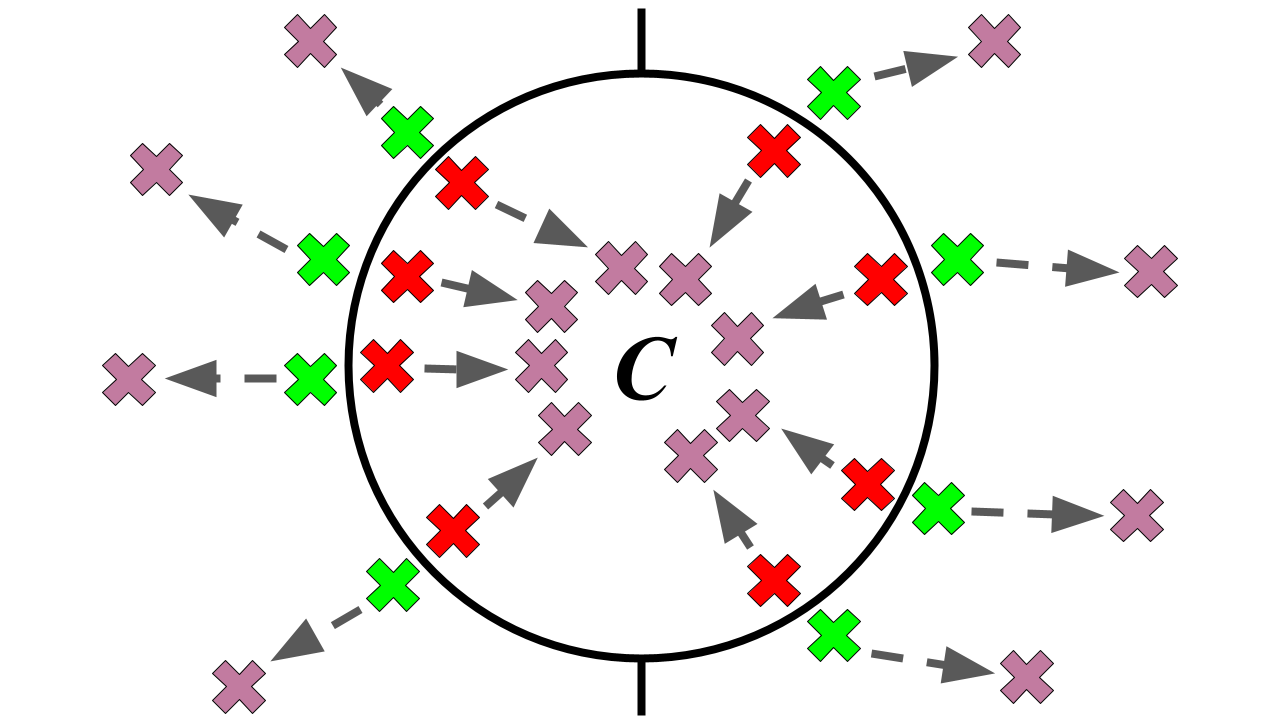}
\caption{}
\end{subfigure}
\end{center}
\vspace{-0.2in}
\caption{Decision boundary exploitation adversarial attack in the teacher feature space for an arbitrary class $c$. The symbols \textcolor{blue}{$\times$}, \textcolor{red}{$\times$}, \textcolor{green}{$\times$}, and \textcolor{violet}{$\times$} represent the original synthetic examples, ``post"-success examples, ``pre"-success examples, and ``deeper" examples, respectively. Best viewed in color.}
\label{fig:adversarial_attack}
\vspace{-0.14in}
\end{figure}

Beyond just identifying the decision boundary in the teacher model, we find that adding an additional attack, which targets the pair of boundary examples to be perturbed ``deeper" into the decision space of their respective classes, leads to additional performance gains. This attack simply clones the boundary examples and performs a targeted BIM attack, where the target labels are the examples' teacher predictions. We find that a single attack step with as little as $\epsilon=1$ is sufficient, with no noticeable benefit to employing a larger $\epsilon$ or more iterations. The full adversarial perturbation process is shown in Fig.~\ref{fig:adversarial_attack} for an arbitrary class $c$. 

% (a) The original synthetic examples, (b) The ``post"-success adversarial examples, (c) The ``pre"-success adversarial examples, and (d) The ``deeper" adversarial examples. 

With a curated synthetic dataset ready to be used for knowledge distillation on a specific teacher network, and a method for altering the data to identify the decision boundaries in the teacher, we will next discuss our approach for distilling knowledge from the teacher to the student. 

\subsection{Knowledge Distillation} \label{sec:distillation}

In this section, we describe our method for transferring knowledge from teacher to student, which is based upon the standard approach for KD \cite{Hinton2015a}. Given a pretrained teacher network $\mathcal{F}_t$ and a randomly initialized student network $\mathcal{F}_s$, an example is first augmented then perturbed using $\mathcal{F}_t$. After completing the adversarial attack process, the newly perturbed example is passed forward through both networks to produce the teacher and student softmax distributions $p_t$ and $p_s$, respectively. The knowledge distillation loss is computed on these softmax scores as
\begin{align}
    \mathcal{L}(p_t\ ||\ p_s) &= \sum_{i\ \in\ \mathcal{C}} \left[\ p_{t(i)} log\ p_{t(i)} - p_{t(i)} log\ p_{s(i)}\ \right]
\end{align}

\noindent which is simply the KL-divergence. Like \cite{Hinton2015a}, we also include a temperature parameter $\tau$ to adjust the entropy of the output softmax distributions from the teacher and student networks before they are used to compute the loss (\ie, $p_t \propto exp(\frac{log\ p_t}{\tau})$ and $p_s \propto exp(\frac{log\ p_s}{\tau})$).

We find that by combining our teacher-specific synthetic dataset with heavy data augmentation and our proposed adversarial attacks, knowledge can be sufficiently transferred from a teacher to a student using the standard KD approach.

\section{Experiments} \label{sec:experiments}

% To evaluate our proposed approach, we conducted a series of KD experiments using several different classification datasets and network architectures, compared to relevant existing DFKD approaches, and examined the various components of our method. We conclude our experimentation with a privacy study.
We first conducted experiments in a self-distillation setting, followed by a typical large teacher to small student KD scenario. Next, we compared to other relevant DFKD approaches, and then investigated the various components of our approach via ablation and sensitivity studies. Lastly, we conducted a qualitative privacy study on various models.

\smallskip
\noindent \textbf{Datasets and Networks.} We employed three established image classification datasets for our evaluation: CIFAR10 (C10) \cite{Krizhevsky2009a}, CIFAR100 (C100) \cite{Krizhevsky2009a}, and Tiny ImageNet (Tiny ImgNet) \cite{Deng2009a}, which contain 10, 100, and 200 classes, respectively. For all datasets, we randomly sampled 10\% of the training examples class-wise for validation. The remaining training examples were used to train the teacher networks, with the validation set used to select the best model. 

% We employed three established image classification datasets for our evaluation. CIFAR10 (C10) \cite{Krizhevsky2009a} contains 10 classes, each with 5K and 1K images for training and testing, respectively. The larger, yet distinct, CIFAR100 (C100) \cite{Krizhevsky2009a} has 100 classes, each with 500 training and 100 testing examples. Finally, Tiny ImageNet (Tiny ImgNet) \cite{Deng2009a} is comprised of 100K training and 10K validation (used as testing) images distributed across 200 classes. For all datasets, we randomly sampled 10\% of the training examples class-wise for validation. The remaining training examples were used to train the teacher networks with the validation set used to select the best model. 

A variety of network architectures are considered for each of the datasets, consisting of ResNets (RN) \cite{He2016a}, Wide ResNets (WRN) \cite{Zagoruyko2016a}, ResNeXts (RX) \cite{Xie2017a}, MobileNetV2 (MNv2) \cite{Sandler2018a}, and ShuffleNetV2 (SNv2) \cite{Ma2018a}. Specific variants for each dataset will be discussed in the appropriate sections below. We trained each selected network and dataset combination using cross entropy to establish a baseline for each model and report those results in the tables. 

\smallskip
\noindent \textbf{Baseline Training Details.} To train our teachers and reference baseline models, we used SGD with momentum (0.9) and weight decay (1e-4) and a one-hot cross-entropy loss. All networks were trained for 400 epochs with a batch size of 256 and a half-period cosine learning rate scheduler, beginning with an initial value of 0.1. Data augmentation consisted of RandAugment \cite{Cubuk2020a} ($n=2, m=14$), random horizontal flipping, and random cropping with padding. The epoch yielding the best validation accuracy was used to select the final model.
% For each of the datasets, we also trained a ResNeXt50-32x8 network with stronger regularization in the form of mixup ($\alpha = 0.2$) \cite{Zhang2018a}.

\smallskip
\noindent \textbf{Distillation Training Details.} For our DFKD experiments, we similarly employed SGD with momentum (0.9) and weight decay (1e-4) for our student networks, and utilized the standard KD loss ($\tau = 20$) as defined previously \cite{Hinton2015a}. Distillation was performed for 400 epochs for all datasets using a half-period cosine learning rate scheduler with an initial value of 0.1, and batch sizes of 128 for CIFAR10/100 and 64 for Tiny ImageNet. Our synthetic datasets consisted of 25K OpenGL shader images for each of the datasets. We additionally experimented with increasing the number of synthetic examples used on CIFAR10 in our sensitivity studies. Our standard data augmentation regime for KD consisted of RandAugment ($n=4, m=14$), random elastic transform, random inversion, random horizontal flipping, and random cropping with padding. Furthermore, we also included mixup \cite{Zhang2018a} with $\alpha \in \mathcal{U}(0, 1)$, similar to \cite{Beyer2022a}. As previously stated, we included more augmentation for KD to add a large amount of data diversity into the process.

The last major component of our approach is the adversarial attacks. The first (``border") attack for identifying decision boundaries in the teacher network is a targeted $L_\infty$ BIM attack with $\epsilon = 10$, an initial step size of $\alpha=1$, and a maximum of 12 iterations. The softmax threshold for this first attack is $0.95$, with other values considered later in our sensitivity studies.  After identifying the decision boundary and removing examples that did not meet the filtering criteria (Sect.~\ref{sec:adversarial}), the second (``deeper") adversarial attack is then applied, which is a targeted $L_\infty$ BIM attack with $\epsilon = \alpha = 1$ for a single step (no softmax threshold). % Our approach was implemented using PyTorch \cite{Paszke2019a}. %and all models were trained and evaluated on a single NVIDIA A100 GPU, except four A100s for ImageNet.

\smallskip
\noindent \textbf{Evaluation.} Three separate runs were performed for each distillation experiment using sequential seed values of 1, 2, and 3, which were made more complex using MD5 as suggested in \cite{Jones2010a, Picard2021a, Davis2022a}. The mean and standard deviation of the three trials are reported. The student model from the last epoch of distillation was selected for evaluation as validation checking \textit{cannot} be performed in this \textit{data-free} context. 

\subsection{Results}

The experimental results are shown in Table \ref{tab:self_distillation} for self-distillation, Table \ref{tab:distilling_down} for distilling from large teacher to smaller student, Table \ref{tab:comparisons} for comparisons to related approaches, and Table \ref{tab:ablation_and_sensitivity} for ablation and sensitivity studies. Our qualitative privacy study is presented in Sect.~\ref{sec:privacy}.

\smallskip
\noindent \textbf{Self-Distillation.} To evaluate how well a network can transfer its knowledge to another network that is the same architecture, we chose one smaller network and one larger network for each dataset. This consisted of MNv2 and WRN22x8 for CIFAR10, RN18 and RX29-64x4 for CIFAR100, and RN18 and RX29-64x8 for Tiny ImageNet. Results of this experiment are shown in Table \ref{tab:self_distillation}. We see that for all datasets and network architectures, knowledge can be transferred from a pretrained network to a freshly initialized one using our approach, albeit at a small performance loss (which is typical for DFKD). Our method still achieves competitive performance levels, typically falling within 2\% of the network's baseline accuracy. The accuracy degradation seen in our distilled students can be lessened if we include more synthetic examples or train for more epochs, as will be shown later. Furthermore, we will also revisit these self-distillation models later in our privacy studies. 

%%% Self-distillation
\begin{table}[t]
\scriptsize
\setlength\extrarowheight{0.5pt}
\begin{center}
\begin{tabular}{c || c || P{1cm} || P{1.5cm} }
\hline
& & \multicolumn{2}{c}{Accuracy} \\
Dataset         & Model & T / S & T $\rightarrow$ S \\ \hhline{=||=||=||=}
\multirow{2}{*}{C10}             &   MNv2   & 95.05  &  \val{93.86}{0.10} \\
                &   WRN22x8   &  96.21 & \val{95.18}{0.09} \\ \hline
\multirow{2}{*}{C100}            &   RN18   & 74.30 & \val{72.56}{0.23} \\
                &   RX29-64x4   & 77.82 & \val{76.67}{0.04}  \\ \hline
\multirow{2}{*}{\shortstack[c]{Tiny\\ImgNet}}            &   RN18   & 61.81 & \val{60.59}{0.08} \\
                &   RX29-64x8   & 64.16 & \val{62.09}{0.24}\\ \hline
\end{tabular}
\vspace{-0.1cm}
\caption{Accuracy on various datasets when self-distilling from two networks of the same architecture using our approach.}
\label{tab:self_distillation}
\end{center}
\vspace{-0.3cm}
\end{table}

\smallskip
\noindent \textbf{Distilling Down.} In most cases, KD is performed between a large, pretrained teacher model and a much smaller, freshly initialized student. Here, we employed teacher$\rightarrow$student pairs of RX29-64x4$\rightarrow$MNv2 and WRN22x8$\rightarrow$RN18 for CIFAR10, RN50$\rightarrow$MNv2 and RX29-64x4$\rightarrow$RN18 for CIFAR100, and WRN28x10$\rightarrow$SNv2 and RX29-64x8$\rightarrow$RN18 for Tiny ImageNet. We additionally trained a teacher RX50-32x8 model with mixup ($\alpha = 0.2$) and distilled it to a RN18 for each of the datasets to investigate the behavior of our approach when an even stronger teacher is employed. In the future, we will include one teacher$\rightarrow$student pair for the standard ImageNet dataset. This experiment will use BiT-M-ResNet 152x2 \cite{Kolesnikov2020a} as a teacher and RN50 as a student, similar to \cite{Beyer2022a}, to demonstrate how our approach can scale up to larger datasets. This network pair was trained for 600 epochs with 50K synthetic OpenGL images. Pretrained weights for that teacher model were taken from \cite{Beyer2022a}. Results for all four datasets are reported in Table \ref{tab:distilling_down}.

Knowledge is still able to be transferred from the large teacher to the small student using our proposed method. However, we observe that there is a larger gap between the baseline cross entropy student and the distilled student compared to the self-distillation results. It is likely that, because of either the change in network architecture or the complexity of the larger teacher models, more synthetic examples, even stronger data augmentation, or more distillation epochs could be needed to narrow the gap in performance.

%%% Distilling down
\begin{table}[t]
\scriptsize
\setlength\extrarowheight{0.5pt}
\begin{center}
\begin{tabular}{c || c | c || P{0.7cm} | P{0.7cm} || P{1.1cm} }
\hline
& & & \multicolumn{3}{c}{Accuracy} \\
Dataset  & Teacher & Student & T & S & T $\rightarrow$ S \\ \hhline{=||=|=||=|=||=}
\multirow{3}{*}{C10}        &  RX29-64x4  &  MNv2  & 95.98 & 95.05 & \val{91.89}{0.15}  \\
                            &  WRN22x8  &  RN18  & 96.21 & 95.23 & \val{94.12}{0.05}  \\
                            &  RX50-32x8  &  RN18  & 96.90 & 95.23 & \val{93.93}{0.14} \\ \hline
\multirow{3}{*}{C100}        &  RN50  &  MNv2  & 77.44 & 76.15 & \val{71.99}{0.13} \\
                &  RX29-64x4  &  RN18  & 77.82 & 74.30 & \val{71.73}{0.13} \\
                &  RX50-32x8  &  RN18  & 81.54 & 74.30 & \val{72.92}{0.18} \\ \hline
\multirow{3}{*}{\shortstack[c]{Tiny\\ImgNet}}            &  WRN28x10  &  SNv2  & 65.13 & 63.33 & \val{59.10}{0.20} \\
                &  RX29-64x8  &  RN18  & 64.16 & 61.81 &  \val{56.05}{0.22} \\
                &  RX50-32x8  &  RN18  & 71.02 & 61.81 & \val{54.37}{0.33} \\ \hline
ImgNet          &  B-RN152x2  &  RN50  & 85.15 & 80.86 & TBD \\ \hline
\end{tabular}
\vspace{-0.1cm}
\caption{Accuracy on various datasets when distilling from a large teacher to a smaller student using our approach.}
\label{tab:distilling_down}
\end{center}
\vspace{-0.8cm}
\end{table}

\smallskip
\noindent \textbf{Comparisons.} Next, we compared to the existing DFKD methods of Contrastive Model Inversion (CMI) \cite{Fang2021a} and Pseudo Replay Enhanced DFKD (PRE) \cite{Binici2022a}. We include the reported RN34$\rightarrow$RN18 results of CMI and PRE from \cite{Binici2022a}, which used their own teacher models and optimal hyperparameter settings (denoted as CMI* and PRE* in the table). However, for a more direct comparison, we also ran the code for CMI/PRE with the same hyperparameter settings and report those scores for our own teacher$\rightarrow$student pairs as a dual-comparison (denoted as CMI and PRE in the table). When implementing these approaches, we used \textit{their code} available on GitHub and the hyperparameters they specify for these datasets. Moreover, the authors only provide one set of hyperparameter values which we used for all experiments. We emphasize that we \textit{do not alter the code in any form} besides simply replacing their teachers with ours and removing validation-based checkpointing provided in their code (as this is \textit{data-free} KD where there is no ground truth imagery). In the case of Tiny ImageNet, we did need to decrease the batch size due to memory, however we proportionally increased the number of iterations for each epoch to account for this. Besides adjusting the batch size in this one situation, all hyperparameter settings are \textit{as specified} in the respective works. 

Comparisons of CMI, PRE, and our method for CIFAR10, CIFAR100, and Tiny ImageNet are shown in Table \ref{tab:comparisons}, with the best scores emphasized in \textbf{bold}. We examined scenarios of self-distillation from teacher to a freshly initialized teacher and distilling from teacher to a smaller, freshly initialized student. We used WRN22x8, RX29-64x4, and RX29-64x8 as the teacher models for CIFAR10, CIFAR100, and Tiny ImageNet, respectively, and RN18 as the student network for all three datasets. Thus, our teacher networks are different from those used in the original works for CMI and PRE, but the student network is exactly the same. As mentioned, we do include the results for CMI and PRE reported in \cite{Binici2022a} (including their reported baseline network scores) which used a RN34$\rightarrow$RN18 network pair for all three datasets. Note that \cite{Binici2022a} also observed difficulties employing CMI to Tiny ImageNet, thus they did not provide any scores for that dataset.

As can be seen in Table \ref{tab:comparisons}, our approach significantly outperforms CMI and PRE on all datasets and distillation scenarios for the values reported in \cite{Binici2022a} and our obtained scores. It could be argued that we did not search for the optimal hyperparameter values for the other approaches with respect to our teacher networks. However, we reiterate that we used the settings provided by the authors, indicating that substantial hyperparameter tuning is always required when employing different teacher networks \cite{Fang2021b}. Our approach on the other hand requires very minimal amounts of hyperparameter tuning and as will be shown in the next set of experiments, our method is much more robust to the choice of settings. Furthermore, the scores we obtained with our approach have significantly less variability compared to CMI and PRE, as shown by the standard deviation of the results.

%%% Comparison
\begin{table}[t]
\scriptsize
\setlength\extrarowheight{0.5pt}
\begin{center}
\begin{tabular}{c || P{0.8cm} || P{0.55cm} | P{0.55cm} || P{1.05cm} | P{1.05cm} }
\hline
& & \multicolumn{4}{c}{Accuracy} \\
Dataset  & Method & T & S & T$_A$$\rightarrow$S$_A$ & T$_A$$\rightarrow$S$_B$ \\ 
\hhline{=||=||=|=||=|=}
\multirow{5}{*}{\shortstack[c]{C10\\A: WRN22x8\\B: RN18}}
                &  \ \ CMI*  &    \tikzmarknode{b3}{95.40}  & \tikzmarknode{b4}{95.20} & - & \val{82.40}{4.07}  \\
                &  \ \ PRE*  &    \tikzmarknode{e3}{}  & \tikzmarknode{e4}{} & - & \val{87.40}{3.21} \\ \cline{2-6}
                &  CMI       &    \tikzmarknode{b1}{96.21}  & \tikzmarknode{b2}{95.23} & \val{76.74}{3.47} & \val{64.69}{3.68}  \\
                &  PRE       &                              &  & \val{87.69}{1.30} & \val{60.67}{16.7} \\ \cline{2-6}
                & Ours       &    \tikzmark{e1}             & \tikzmark{e2} & \bval{95.18}{0.09} & \bval{94.12}{0.05} \\ \hline
\multirow{5}{*}{\shortstack[c]{C100\\A: RX29-64x4\\B: RN18}} 
                &  \ \ CMI*  &    \tikzmarknode{b5}{77.90}  & \tikzmarknode{b6}{77.1} & - & \val{55.20}{4.91}  \\
                &  \ \ PRE*  &    \tikzmarknode{e5}{}  & \tikzmarknode{e6}{} & - & \val{70.20}{3.33}  \\ \cline{2-6}
                &  CMI  &    \tikzmarknode{b7}{77.82}  & \tikzmarknode{b8}{74.30} & \val{54.63}{0.34} & \val{46.60}{0.73}  \\
                &  PRE  &      &  & \val{56.41}{3.62} & \val{41.39}{4.44}  \\ \cline{2-6}
                & Ours & \tikzmarknode{e7}{}  & \tikzmarknode{e8}{} & \bval{76.67}{0.04} & \bval{71.73}{0.13} \\ \hline 
\multirow{5}{*}{\shortstack[c]{Tiny ImgNet\\A: RX29-64x8\\B: RN18}}            
                &  \ \ CMI*  &  \tikzmarknode{b9}{71.20}  & \tikzmarknode{b10}{64.90} & - & -  \\
                &  \ \ PRE*  &  \tikzmarknode{e9}{}  & \tikzmarknode{e10}{} & - & \val{46.30}{3.32}  \\ \cline{2-6}
                &  CMI  &  \tikzmarknode{b11}{64.16}  & \tikzmarknode{b12}{61.81} & \val{33.80}{0.24} & \val{30.81}{0.50}  \\
                &  PRE  &    &  & \val{9.14}{3.75} & \val{8.59}{1.52}  \\ \cline{2-6}
                & Ours & \tikzmarknode{e11}{} & \tikzmarknode{e12}{} & \bval{62.09}{0.24} & \bval{56.05}{0.22} \\ \hline
\end{tabular}
\begin{tikzpicture}[overlay, remember picture]
\draw[shorten <=0.1cm, ->,black,thick] (pic cs:b1) -- (pic cs:e1);
\draw[shorten <=0.1cm, ->,black,thick] (pic cs:b2) -- (pic cs:e2);
\draw[shorten <=0.1cm, ->,black,thick] (pic cs:b3) -- (pic cs:e3);
\draw[shorten <=0.1cm, ->,black,thick] (pic cs:b4) -- (pic cs:e4);
\draw[shorten <=0.1cm, ->,black,thick] (pic cs:b5) -- (pic cs:e5);
\draw[shorten <=0.1cm, ->,black,thick] (pic cs:b6) -- (pic cs:e6);
\draw[shorten <=0.1cm, ->,black,thick] (pic cs:b7) -- (pic cs:e7);
\draw[shorten <=0.1cm, ->,black,thick] (pic cs:b8) -- (pic cs:e8);
\draw[shorten <=0.1cm, ->,black,thick] (pic cs:b9) -- (pic cs:e9);
\draw[shorten <=0.1cm, ->,black,thick] (pic cs:b10) -- (pic cs:e10);
\draw[shorten <=0.1cm, ->,black,thick] (pic cs:b11) -- (pic cs:e11);
\draw[shorten <=0.1cm, ->,black,thick] (pic cs:b12) -- (pic cs:e12);
\end{tikzpicture}
\vspace{-0.1cm}
\caption{Comparison of related DFKD approaches on CIFAR10, CIFAR100, and Tiny ImageNet. Results on RN34$\rightarrow$RN18 obtained from \cite{Binici2022a} are denoted as CMI* and PRE*.}
\label{tab:comparisons}
\end{center}
\vspace{-0.85cm}
\end{table}

\smallskip
\noindent \textbf{Ablation \& Sensitivity Studies.} In this section, we conducted three different studies: an ablation study on our adversarial attack, a sensitivity study on the attack choices and settings, and a sensitivity study on the dataset and training settings. For these experiments, we specifically focused on CIFAR10 with a WRN22x8 teacher and a RN18 student. Results of all three studies are presented in Table \ref{tab:ablation_and_sensitivity}. 

For the attack ablation study, we observed that even when both adversarial attacks (``border" and ``deeper") are \textit{completely removed} (\ie, just performing KD with the synthetic examples and data augmentation), information can still be transferred from the teacher to the student in a data-free manner. This suggests that employing in-domain versus out-of-domain data for standard KD does not necessarily matter as long as the dataset uniformly samples the teacher's output labels. Thus, in the case of the out-of-domain experiment in \cite{Beyer2022a}, it is likely that the base out-of-domain data predictions in the teacher model were long-tail imbalanced, which did not sample the decision space in the teacher sufficiently for performing KD. Furthermore, we see that our approach requires both the first border attack and the second deeper attack to reach the best performance possible. Additionally, the inclusion of the ``pre"-success examples in the first border attack results in greater performance than using the ``post"-success examples alone. However, filtering the examples based on the conditions outlined in Sect.~\ref{sec:adversarial} does not produce significantly different results from our full approach.

%%% Ablation Studies
\begin{table}[t]
\scriptsize
\setlength\extrarowheight{0.5pt}
\begin{center}
\begin{tabular}{P{1.5cm}|P{3.5cm} || P{1.1cm} }
\hline
Study & Experiment & T $\rightarrow$ S \\ 
\hhline{==||=}
- & \multicolumn{1}{l||}{Full Approach}   & \val{94.12}{0.05} \\ \hline
\multirow{5}{*}{Ablation} & \multicolumn{1}{r||}{($-$) Both Adversarial Attacks} & \val{93.07}{0.27} \\ 
& \multicolumn{1}{r||}{($-$) ``Border" First Attack Examples} & \val{93.38}{0.15} \\ 
& \multicolumn{1}{r||}{($-$) ``Deeper" Second Attack Examples} & \val{93.01}{0.11} \\ 
& \multicolumn{1}{r||}{($-$) ``Pre"-Success Examples} & \val{93.10}{0.17} \\ 
& \multicolumn{1}{r||}{($-$) Filter Conditions} & \val{94.19}{0.19} \\
\hhline{==||=}
\multirow{7}{*}{\shortstack[c]{Attack\\Sensitivity}} & \multicolumn{1}{r||}{PGD-based Attack} & \val{91.62}{0.70} \\  \cline{2-3}
& \multicolumn{1}{r||}{$\epsilon=6$ Initial Attack Epsilon} & \val{94.15}{0.17} \\ 
 & \multicolumn{1}{r||}{$\epsilon=14$ Initial Attack Epsilon} & \val{94.12}{0.22} \\ 
 & \multicolumn{1}{r||}{$\epsilon=4$ Second Attack Epsilon} & \val{94.23}{0.16} \\ \cline{2-3}
& \multicolumn{1}{r||}{No Softmax Threshold} & \val{94.04}{0.15} \\ 
& \multicolumn{1}{r||}{0.50 Softmax Threshold} & \val{93.92}{0.18} \\ 
& \multicolumn{1}{r||}{0.75 Softmax Threshold} & \val{94.11}{0.13} \\ 
\hhline{==||=}
\multirow{12}{*}{\shortstack[c]{Dataset\\and\\Training\\Sensitivity}} & \multicolumn{1}{r||}{Minimal Data Augmentation w/ mixup} & \val{93.57}{0.19} \\ 
& \multicolumn{1}{r||}{No Data Augmentation w/ Mixup} & \val{93.03}{0.14} \\ 
& \multicolumn{1}{r||}{Standard Data Augmentation w/o Mixup} & \val{92.92}{1.08} \\ 
& \multicolumn{1}{r||}{Minimal Data Augmentation w/o Mixup} & \val{89.40}{0.67} \\ 
& \multicolumn{1}{r||}{No Data Augmentation w/o Mixup} & \val{34.67}{5.11} \\ \cline{2-3}
& \multicolumn{1}{r||}{Attack with Normal \& Mixup} & \val{94.32}{0.31} \\ 
& \multicolumn{1}{r||}{Distill with Normal, Mixup \& Adversarial} & \val{94.37}{0.05} \\ \cline{2-3}
& \multicolumn{1}{r||}{500 Examples per Class} & \val{62.83}{35.0} \\ 
& \multicolumn{1}{r||}{1000 Examples per Class} & \val{90.93}{0.60} \\ 
& \multicolumn{1}{r||}{5000 Examples per Class} & \val{94.59}{0.21} \\ \cline{2-3}
& \multicolumn{1}{r||}{200 Distillation Epochs} & \val{92.32}{0.72} \\ 
& \multicolumn{1}{r||}{800 Distillation Epochs} & \val{94.61}{0.11} \\ \hline
\end{tabular}
\vspace{-0.1cm}
\caption{Ablation and sensitivity studies on CIFAR10 with a WRN22x8 teacher and RN18 student.}
\label{tab:ablation_and_sensitivity}
\end{center}
\vspace{-0.85cm}
\end{table}

Next, we investigated the impact of our choices for our adversarial attacks. We altered the first border attack to be a Projected Gradient Descent (PGD) attack (\ie, a random uniform perturbation of $\epsilon=10$ was applied to each example before the attack was conducted) and saw much worse results than using BIM. We observed that the initial perturbation of PGD tends to push most examples toward being initially classified as one of a few classes. Thus, likely leading our attack to mostly identify the decision boundary between these few labels and the targets. As for the values of $\epsilon$ and the softmax threshold of the first border attack (the only hyperparameters for our method), we see that our approach is fairly robust to the values for these hyperparameters. In comparison to generator-based methods (Table \ref{tab:comparisons}), we see that our approach is much less sensitive to settings.

To conclude our ablation and sensitivity studies, we examined the sensitivity of our approach with respect to how much input data augmentation is utilized, which examples are used in the adversarial attacks and in the distillation loss, how many synthetic examples are used during distillation, and how long the student network is trained. For the data augmentations, we experimented with three different levels of data augmentations: standard (Sect.~\ref{sec:experiments} Distillation Training Details), minimal (no RandAugment, random elastic transform, or random inversion), and none. Additionally, we investigated the effects of removing mixup from our approach. From Table \ref{tab:ablation_and_sensitivity}, it is obvious to see how crucial data augmentation and mixup are to our method. When no data augmentation is applied, mixup alone enables our approach to achieve over 90\% accuracy on CIFAR10, which is about 60\% higher than no data augmentation and no mixup. Conversely, when standard data augmentation is used but mixup is not, we see similar performance. Combining strong amounts of data augmentation and mixup together helps achieve the best performance possible. 

Next, we examined the effects of using more types of examples in the adversarial attacks and in the distillation loss. By including the original data augmented synthetic examples (pre-mixup) with the mixup examples in the adversarial attacks, performance gains can be seen. Furthermore, when including both the original data augmented synthetic examples and the mixup examples (pre-attack) with the adversarially perturbed examples in the KD loss computation, performance gains can also be seen. We also observed that sizable performance gains can be obtained by increasing either the number of synthetic examples per class or the number of epochs for performing KD. Through this sensitivity study, we show that, while we obtained state-of-the-art results in our previous experiments, there are several avenues for increasing performance scores even further. Thus, in our CIFAR100 and Tiny ImageNet experiments and in our ``distilling down" experiments, it is likely that we could bridge the gap between our distilled students and the baseline cross entropy student by increasing the number of synthetic examples and conducting more distillation epochs.

\subsection{Privacy} \label{sec:privacy}

\begin{figure}[t]
\begin{center}
\begin{subfigure}{.105\textwidth}
\includegraphics[width=1.7cm]{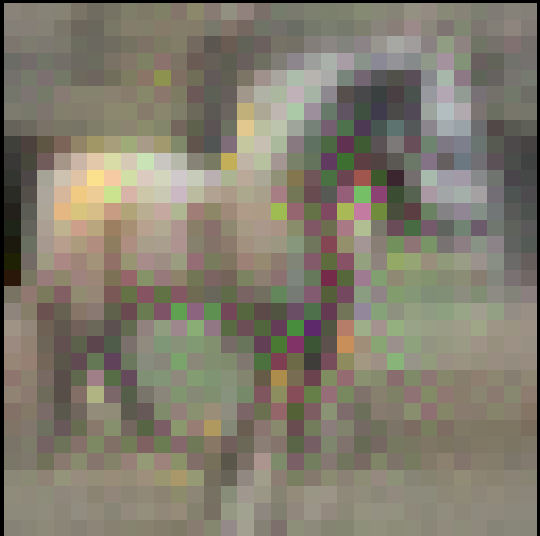}
\end{subfigure}
\begin{subfigure}{.105\textwidth}
\includegraphics[width=1.7cm]{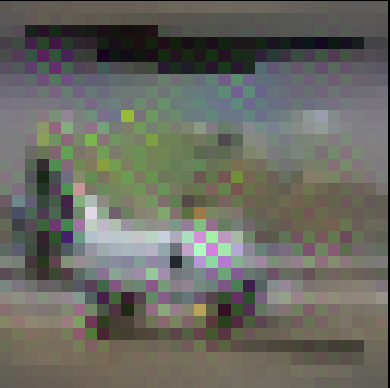}
\end{subfigure}
\begin{subfigure}{.105\textwidth}
\includegraphics[width=1.7cm]{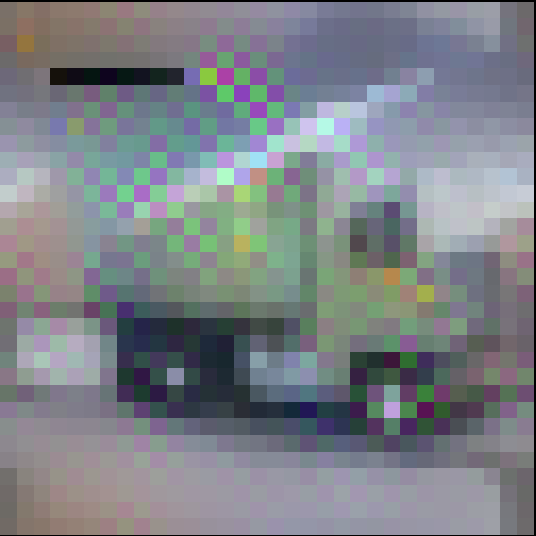}
\end{subfigure}
\begin{subfigure}{.105\textwidth}
\includegraphics[width=1.7cm]{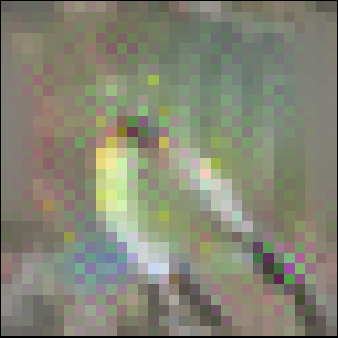}
\end{subfigure}
\\ \vspace{0.1cm} \hrule \vspace{0.1cm}
\begin{subfigure}{.105\textwidth}
\includegraphics[width=1.7cm]{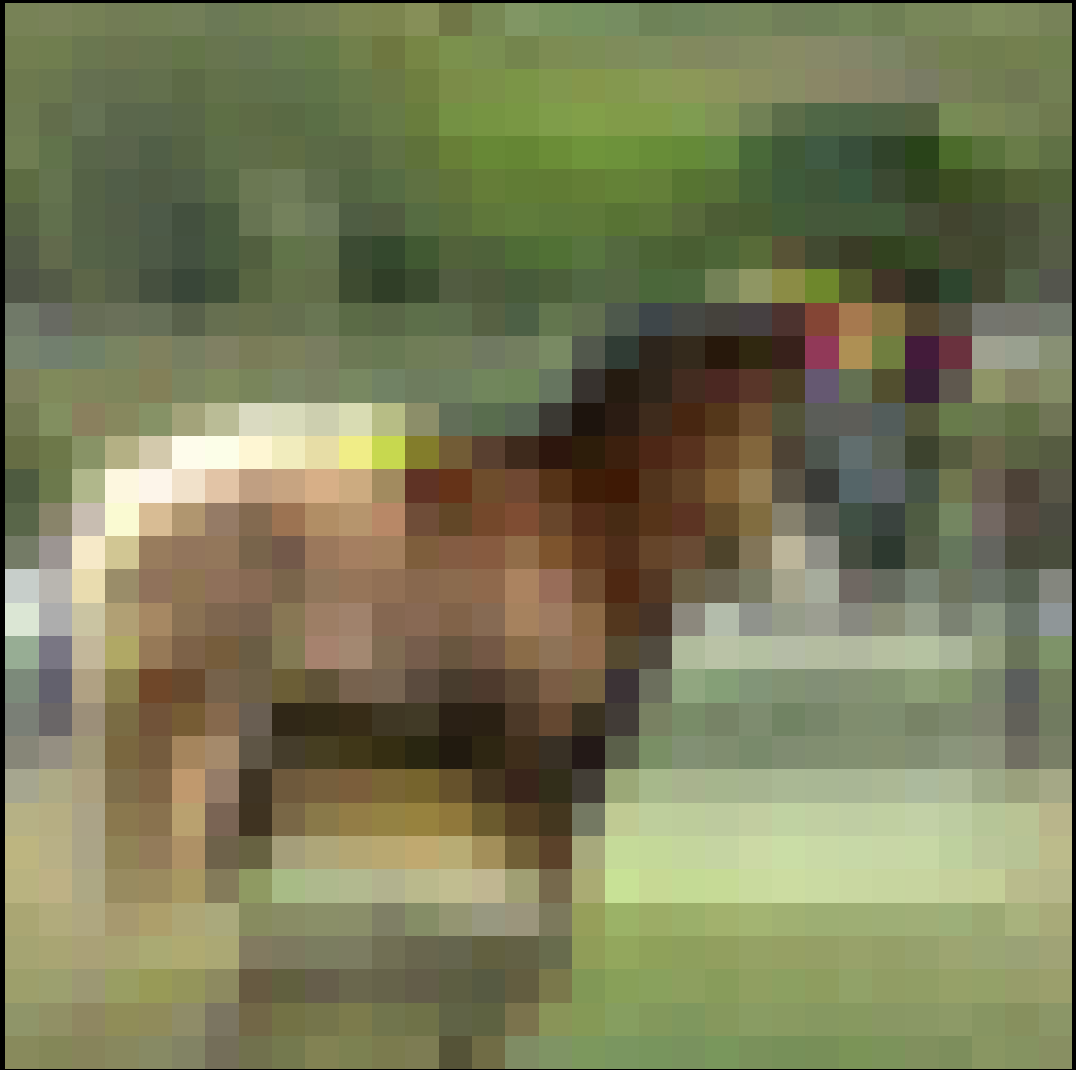}
\end{subfigure}
\begin{subfigure}{.105\textwidth}
\includegraphics[width=1.7cm]{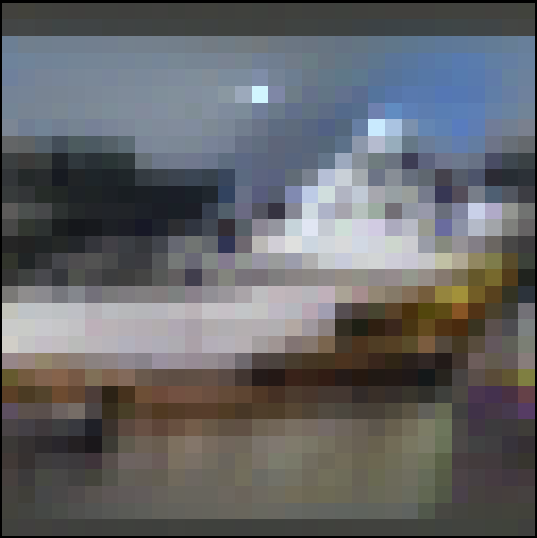}
\end{subfigure}
\begin{subfigure}{.105\textwidth}
\includegraphics[width=1.7cm]{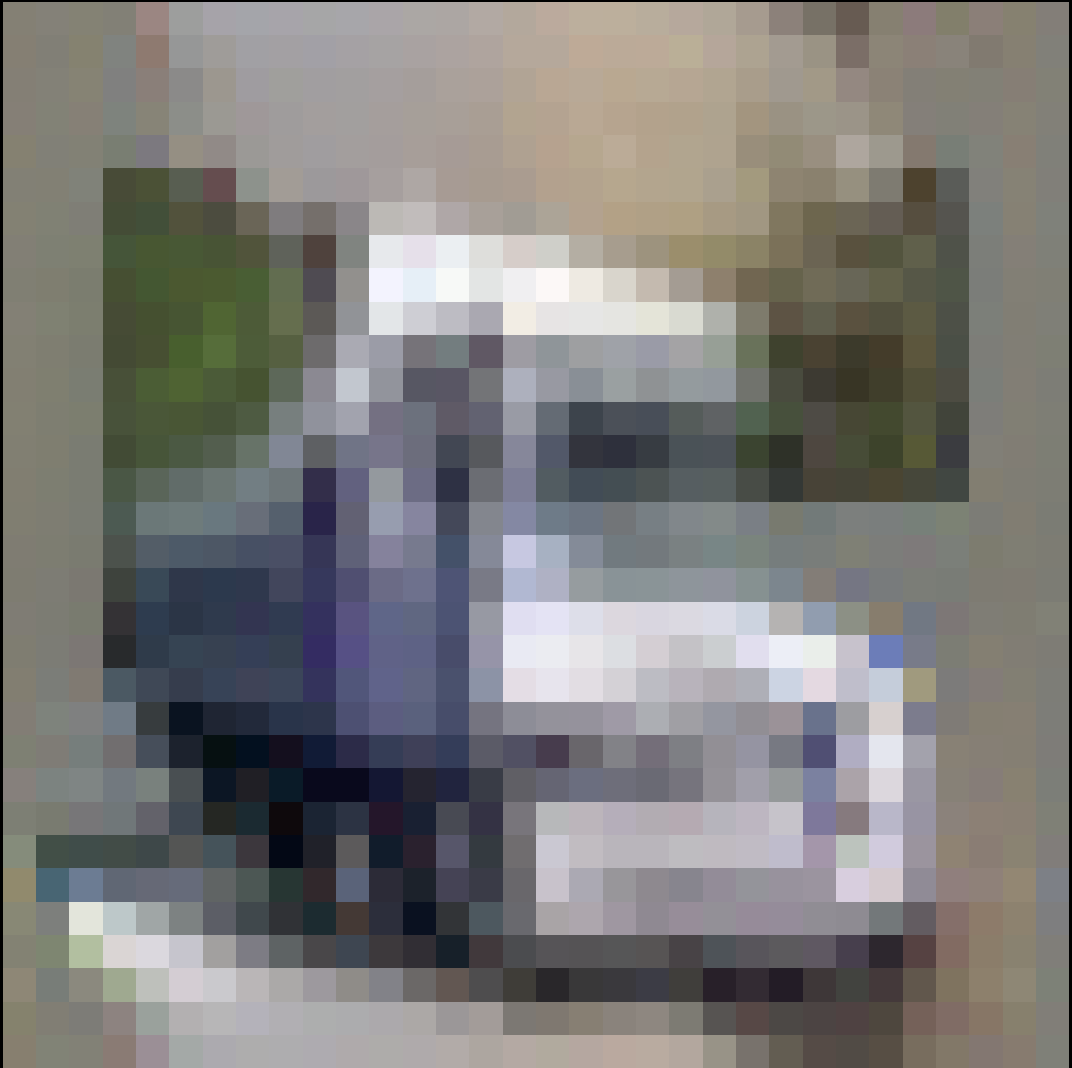}
\end{subfigure}
\begin{subfigure}{.105\textwidth}
\includegraphics[width=1.7cm]{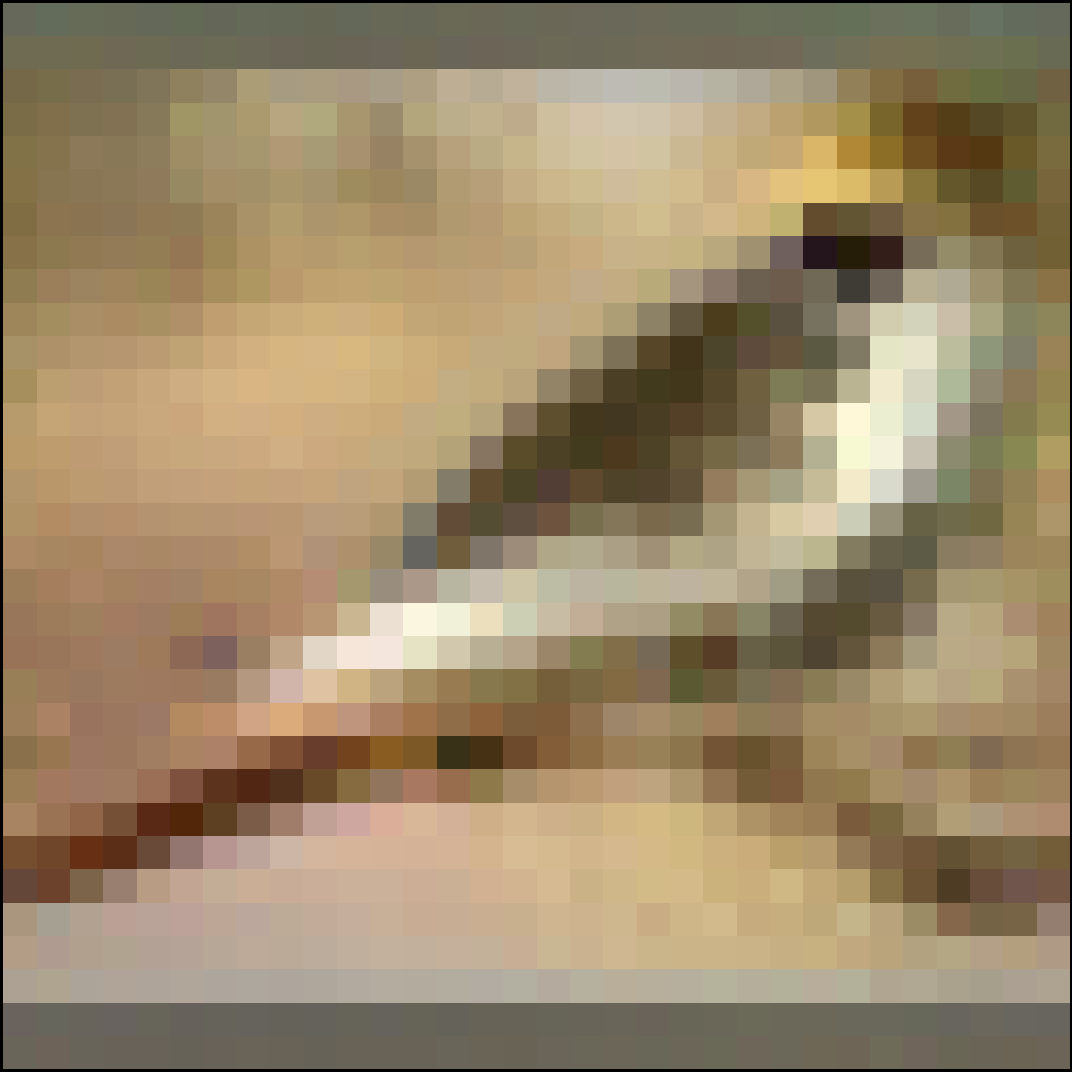}
\end{subfigure}
\\ \vspace{0.1cm} \hrule \vspace{0.1cm}
\begin{subfigure}{.105\textwidth}
\includegraphics[width=1.7cm]{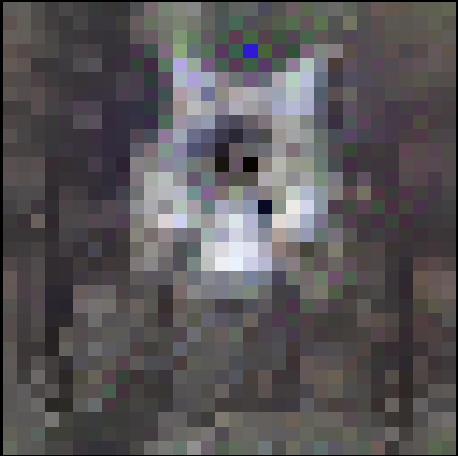}
\end{subfigure}
\begin{subfigure}{.105\textwidth}
\includegraphics[width=1.7cm]{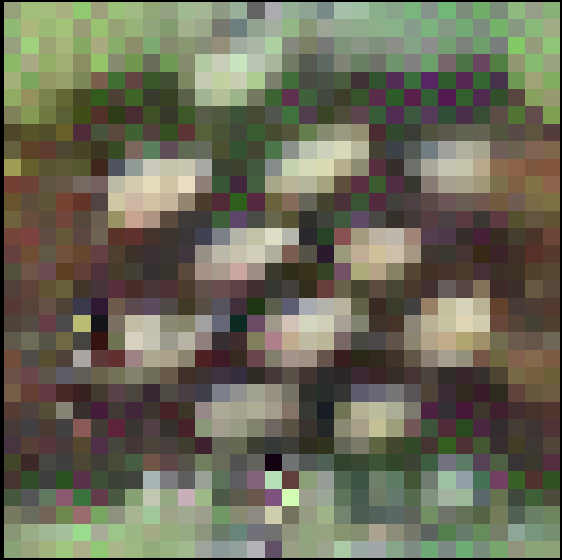}
\end{subfigure}
\begin{subfigure}{.105\textwidth}
\includegraphics[width=1.7cm]{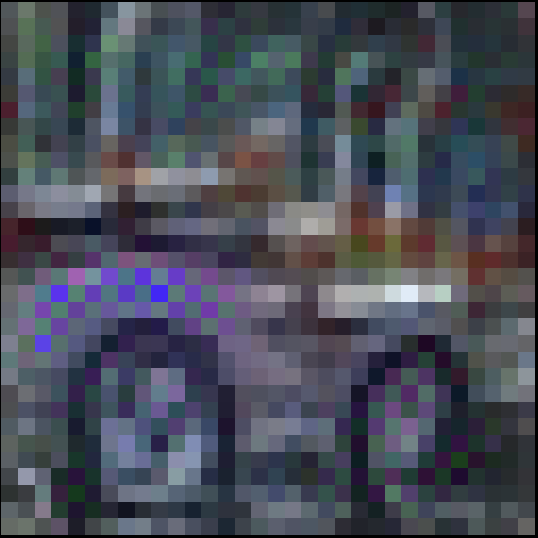}
\end{subfigure}
\begin{subfigure}{.105\textwidth}
\includegraphics[width=1.7cm]{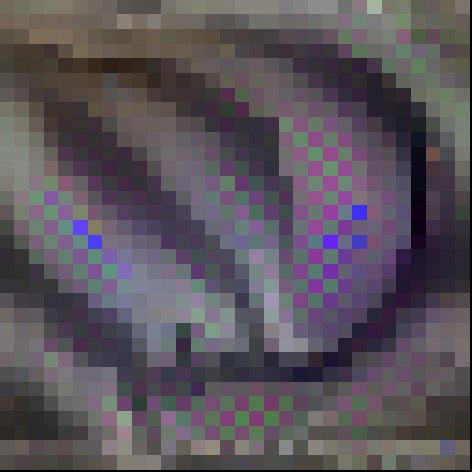}
\end{subfigure}
\end{center}
\vspace{-0.2in}
\caption{Extracted CIFAR10 images from a MobileNetV2 teacher model (top row), a CIFAR10-distilled MobileNetV2 student model (middle row), and a OpenGL-distilled MobileNetV2 student model (bottom row).}
\label{fig:privacy_images}
\vspace{-0.15in}
\end{figure}

Lastly, as recent works have shown that information about the training data can be extracted from various models \cite{Carlini2021a, Carlini2023a, Haim2022a}, we conducted a qualitative privacy examination on our teacher models and distilled students in the context of model inversion and data extraction. For this experiment, we synthesized images using DeepInversion \cite{Yin2020a} applied to three MobileNetV2 models trained on CIFAR10. One is the teacher network that was trained with the original dataset, one is a student network that was self-distilled from the aforementioned teacher using the original CIFAR10 data, and the last is a student network that was self-distilled from the aforementioned teacher using our approach. As mentioned in \cite{Yin2020a}, their approach for model inversion can be sensitive to hyperparameters, thus we examined over 1000 different hyperparameter configurations and selected examples from the configuration yielding the lowest overall loss. These images are presented in Fig.~\ref{fig:privacy_images}, where all images shown were predicted correctly by their respective model. 

It is obvious to see that recognizable images of a horse, airplane, truck, and bird were able to be extracted from the teacher model. Furthermore, we are also able to obtain slightly less detailed, but still recognizable, images of a horse, ship, truck, and bird from the CIFAR10 distilled student. However, when we look at the examples obtained using our synthetic distilled student model, we see that they are largely incoherent. In a couple of cases you can see vague depictions of what could be a cat or a car, but immense detail is missing. For the most part, images extracted from our distilled students appeared to be abstract textures that contained no recognizable objects. Thus, our approach could serve as a method for anonymizing the data that gets naturally embedded into deep neural networks from standard supervised training and ordinary KD.

\section{Conclusion} \label{sec:conclusion}

We proposed a novel approach to data-free knowledge distillation that utilizes procedurally rendered OpenGL shader images, combined with heavy data augmentation and adversarial attacks, to transfer knowledge from a teacher network to a new student model. Our work is straightforward to implement and built upon standard KD, making it extendable to more advanced methods. Experiments demonstrated improved and more stable results over relevant DFKD approaches, establishing new state-of-the-art scores for multiple datasets. Additionally, we showed that our technique can better anonymize the data embedded in the teacher. While we presented results that are superior to other methods, we also see opportunities for improvements (as shown in Table \ref{tab:ablation_and_sensitivity}) which we will explore in the future.

\smallskip
\noindent \textbf{Acknowledgements.} This research was supported by the U.S.\ Air Force Research Laboratory under Contract \#GRT00054740 (Release \#AFRL-2023-5248). We would additionally like to thank Skylar Wurster for his assistance.

%%%%%%%%% REFERENCES
{\small
\bibliographystyle{ieee_fullname}
\bibliography{refs}
}

\end{document}